\documentclass[final,nonatbib]{article}
\usepackage[numbers]{natbib}

\usepackage[final]{neurips_2024_ml4ps}

\usepackage{multirow}

\usepackage{macros_aas}
\usepackage[utf8]{inputenc} 
\usepackage[T1]{fontenc}    
\usepackage{hyperref}       
\usepackage{url}            
\usepackage{booktabs}       
\usepackage{amsfonts}       
\usepackage{nicefrac}       
\usepackage{microtype}      
\usepackage{xcolor}         

\hypersetup{
	colorlinks   = true, 
    urlcolor     = blue, 
    linkcolor    = blue, 
    citecolor    = red   
 }

\usepackage{microtype}
\usepackage{graphicx}
\usepackage{subfigure}
\usepackage{booktabs} 

\usepackage{amsmath}
\usepackage{amssymb}
\usepackage{mathtools}
\usepackage{amsthm}

\usepackage[capitalize,noabbrev]{cleveref}

\usepackage{algorithm}
\usepackage{algorithmic}

\usepackage[textsize=tiny]{todonotes}

\usepackage{typed-checklist}

\theoremstyle{plain}

\theoremstyle{definition}

\theoremstyle{remark}

\newcommand{\authoraffil}[2]{$^{#1}$ \\ #2} 
\newcommand{\assignaffilnumber}[2]{$^#1$#2} 

\newcommand{\nord}{Brian D. Nord}
\newcommand{\ciprijanovic}{Aleksandra \'{C}iprijanovi\'{c}}

\newcommand{\nevin}{Rebecca Nevin}

\newcommand{\nordemail}{\url{nord@fnal.gov}}
\newcommand{\ciprijanovicemail}{\url{aleksand@fnal.gov}}

\newcommand{\nevinemail}{\url{rnevin@fnal.gov}}

\newcommand{\uchicagoAA}{Department of Astronomy and Astrophysics, University of Chicago, Chicago, IL 60637}
\newcommand{\kicp}{Kavli Institute for Cosmological Physics, University of Chicago, Chicago, IL 60637}
\newcommand{\fermilab}{Fermi National Accelerator Laboratory, Batavia, IL 60510}

\newcommand{\ackfundingdeepskies}{We acknowledge the Deep Skies Lab as a community of multi-domain experts and collaborators who’ve facilitated an environment of open discussion, idea generation, and collaboration. This community was important for the development of this project.\\ \\ }

\newcommand{\ackfundingdoegeneral}{Work supported by the Fermi National Accelerator Laboratory, managed and operated by Fermi Research Alliance, LLC under Contract No. DE-AC02-07CH11359 with the U.S. Department of Energy. The U.S. Government retains and the publisher, by accepting the article for publication, acknowledges that the U.S. Government retains a non-exclusive, paid-up, irrevocable, world-wide license to publish or reproduce the published form of this manuscript, or allow others to do so, for U.S. Government purposes.\\ \\ }

\newcommand{\ackfundingldrdnord}{This material is based upon work supported by the Department of Energy under grant No FNAL-LDRD- L2021-004.\\ \\ }

\newcommand{\contribciprijanovic}{
\'{C}iprijanovi\'{c}: Conceptualization, Methodology, Formal analysis, Writing - Review \& Editing, Supervision, Project administration
\\ \\}

\newcommand{\contribnevin}{
Nevin: Conceptualization, Methodology, Formal analysis, Investigation, Writing - Original Draft, Writing - Review \& Editing
\\ \\}

\newcommand{\contribnord}{
Nord: Conceptualization, Methodology, Formal analysis, Resources, Writing - Original Draft, Writing - Review \& Editing, Supervision, Project administration, Funding acquisition
\\ \\}

\author{
\nevin\authoraffil{1}{\nevinemail} \And
\ciprijanovic\authoraffil{1,2}{\ciprijanovicemail} \And
\nord\authoraffil{1,2,3}{\nordemail} \\ \\
\assignaffilnumber{1}{\fermilab}\\
\assignaffilnumber{2}{\uchicagoAA}\\
\assignaffilnumber{3}{\kicp}
}

\definecolor{seagreen}{HTML}{519872}

\title{DeepUQ: Assessing the Aleatoric Uncertainties from two Deep Learning Methods}
\author{
\nevin\authoraffil{1}{\nevinemail} \And
\ciprijanovic\authoraffil{1,2}{\ciprijanovicemail} \And
\nord\authoraffil{1,2,3}{\nordemail} \\ \\
\assignaffilnumber{1}{\fermilab}\\
\assignaffilnumber{2}{\uchicagoAA}\\
\assignaffilnumber{3}{\kicp}
}

\begin{document}

\maketitle

\begin{abstract}

Assessing the quality of aleatoric uncertainty estimates from uncertainty quantification (UQ) deep learning methods is important in scientific contexts, where uncertainty is physically meaningful and important to characterize and interpret exactly.
We systematically compare aleatoric uncertainty measured by two UQ techniques, Deep Ensembles (DE) and Deep Evidential Regression (DER).
Our method focuses on both zero-dimensional (0D) and two-dimensional (2D) data, to explore how the UQ methods function for different data dimensionalities.
We investigate uncertainty injected on the input and output variables and include a method to propagate uncertainty in the case of input uncertainty so that we can compare the predicted aleatoric uncertainty to the known values.
We experiment with three levels of noise.
The aleatoric uncertainty predicted across all models and experiments scales with the injected noise level.
However, the predicted uncertainty is miscalibrated to $\rm{std}(\sigma_{\rm al})$ with the true uncertainty for half of the DE experiments and almost all of the DER experiments. 
The predicted uncertainty is the least accurate for both UQ methods for the 2D input uncertainty experiment and the high-noise level.
While these results do not apply to more complex data, they highlight that further research on post-facto calibration for these methods would be beneficial, particularly for high-noise and high-dimensional settings.

\end{abstract}

\section{Introduction}
\label{sec:introduction}

Physically and statistically interpretable uncertainties are critical for applications in science and industry. 
Uncertainty quantification (UQ) in deep neural networks has gained significant attention, with recent work exploring taxonomies of uncertainties, including domain, epistemic, and aleatoric uncertainties, e.g., \cite{brando2022thesis, brando2023standardizing,gal2022bayesian}. 
Aleatoric uncertainty, $\sigma_{\rm al}$, is significant because, unlike epistemic uncertainty, it is not a result of model limitations but rather an inherent property of the data.
In many cases, aleatoric uncertainty is exactly known because it is produced by a well-understood physical process, allowing us to anticipate not only its expected amplitude but also its distributional characteristics.
For instance, in astrophysics, the Poisson distribution\footnote{The Poisson distribution can be approximated by a Gaussian when the photon rate $\lambda$ is large.} characterizes `shot' or photon noise, while the Normal distribution characterizes read and other forms of thermal or electronic noise.
Developing a coherent framework for benchmarking aleatoric uncertainty estimates from deep networks and assessing their calibration is needed to ensure that the predicted aleatoric uncertainty aligns with our scientific expectations.

UQ methods broadly fall under the categories of Bayesian methods (e.g., Bayesian Neural Networks (BNNs) \cite{LV2000,T2004,PS2017}), Bayesian model averaging (e.g., Deep Ensembles \cite{Lakshminarayanan2016arXiv161201474L}, MC Dropout \cite{GalMCDropout2015arXiv150602142G}), and Evidential Deep Learning \cite{Ulmer2021arXiv211003051U} (e.g., Deep Evidential Regression \cite{Amini2019DER, Meinert2022DER}).
In addition, a class of methods for uncertainty calibration \cite{Nixon2019arXiv190401685N} exist separately in the statistical literature and have recently gained popularity as post-processing tools (e.g., conformal prediction \cite{Angelopoulos2021arXiv210707511A}).
Other work has explored formalized comparison of UQ methods (e.g., \cite{Caldeira2020arXiv200410710C, Scalia2019arXiv191003127S, Bramlageinproceedings, Tran2019arXiv191210066T,Chung2021arXiv210910254C}).
\cite{Scalia2019arXiv191003127S, Tran2019arXiv191210066T} compare aspects of predictive uncertainty distributions, and \cite{Chung2021arXiv210910254C} present an uncertainty toolbox for comparing predictive uncertainties; all of these methods do so without access to true uncertainty values.
Of the few studies testing the exact calibration of predicted uncertainties \cite{Caldeira2020arXiv200410710C, Bramlageinproceedings}, some do not vary data dimensionalities or uncertainty injection types \cite{Caldeira2020arXiv200410710C}, while others vary these factors but do not report mean aleatoric uncertainty or propagate input uncertainty, preventing direct comparison to expected values \cite{Bramlageinproceedings}.

Quantifying how noise on the input variable affects the predictions of aleatoric uncertainty on the output variable from deep learning methods is of critical importance, especially in computer vision, and has not yet received much attention in the literature (e.g., \cite{Rodrigues2023MLS&T...4d5019R, Wright2000}).
The bulk of previous work on aleatoric uncertainty has focused mostly on assessing the predicted aleatoric uncertainty on the output variable $y$ via injecting uncertainty directly on $y$ (for a review, see \cite{Hullermeier2019arXiv191009457H}).
Recently, the statistical field of input uncertainty has intersected with the deep learning literature under the umbrella of UQ (for a review, see \cite{Valdenegro-Toro2024arXiv240618787V}).
Experiments have focused on propagating input uncertainty through a neural network for regression using a Laplace Approximation \cite{Wright2000} as well as through a Taylor series expansion and Monte Carlo sampling approach with a multi-layer perception \cite{Valdenegro-Toro2024arXiv240618787V}.
Assessing input uncertainty is inherently more complex, requiring tractable functional relationships between input and output variables when propagating the uncertainty onto the output variables.

We present a study of regression on tabular (0D) and imaging (2D) data that investigates aleatoric uncertainty for cases where uncertainty is injected on either the input $x$ or the output $y$ variables, providing a more comprehensive understanding of aleatoric uncertainty in regression tasks.
By injecting uncertainty onto the input variable and propagating it to the output variable, we can assess the exact calibration of the predicted uncertainty estimate.
We design a set of desiderata for how the predicted aleatoric uncertainty should behave: i) the predicted uncertainty should scale with the injected uncertainty; ii) the aleatoric uncertainty should be well-calibrated (within $\rm{std}(\sigma_{\rm al})$ of the true uncertainty value); and iii) these desiderata should hold for both data dimensionalities and both uncertainty injection types (input and output).
We do this all for a very simple set of experiments; we caution the reader against applying the conclusions here to all types of data, including real-world datasets.

\section{Deep Learning Methods for Predicting Uncertainty}
\label{sec:dl_uncertainty_methods}

\noindent\textbf{Deep Ensemble:} Ensembling mean-variance estimation networks (MVEs) produces a set of predicted mean and variance values -- `Deep Ensembles' \citep[DE;][]{Lakshminarayanan2016arXiv161201474L}. 
We build upon the DE framework from \cite{Lakshminarayanan2016arXiv161201474L} incorporating several modifications inspired by previous work, including a softplus activation for the $\sigma^2$ output neuron to enforce a positive output value and a $\beta$ modification to the loss function, as introduced in \citep{Seitzer2022arXiv220309168S}.
The modified loss function we use is: $ \mathcal{L}_{\beta-\rm NLL} = 
       \frac{1}{N} \sum_{i=0}^N 
       \left[ \sigma^{2\beta}(x_i)\left[\frac{1}{2}  \mathrm{log}\ \sigma^2(x_i) + 
        \frac{(y_i - \mu(x_i)) ^2}{2\sigma^2(x_i)} 
        + 
        \mathrm{C}\right] \right]$.
The $\beta$-modified loss function helps ensure convergence of the network predictions, avoiding a commonly observed problem in MVEs, where the variance artificially enlarges resulting in a poor estimate of the mean.
We use a $\beta$ value of 0.5, which is recommended by \cite{Seitzer2022arXiv220309168S}, and described in more depth in Appendix \ref{ap:DE}.
The aleatoric uncertainty is the mean of the predicted standard deviations for the ensemble of $K=10$ models: $\sigma_{\rm al} = \sqrt{\frac{1}{K}\sum_{k=1}^K \sigma_k^2}$, where $\sigma_k^2$ is the variance predicted by the  $k$-th network.

\noindent\textbf{Deep Evidential Regression:} Deep Evidential Regression (DER) estimates aleatoric uncertainties via evidential distributions that are directly incorporated into the loss function \citep{Amini2019DER}. 
Instead of requiring an ensemble of networks, it places evidential priors over a Gaussian likelihood function, and the network is trained to learn the hyperparameters of the evidential distribution.

We use the normal-inverse-gamma (NIG) loss from \cite{Meinert2022DER}, which includes an additional term weighted by the width of the $t-$distribution. 
This formulation improves the efficiency and accuracy of training: $\mathcal{L}_{\rm NIG} = \frac{1}{N} \sum_{i=1}^N \left[- {\rm log}\  L_i^{\rm NIG} + \lambda |\frac{y_i-\gamma}{w_{\rm St}}| \Phi \right]$, where $\lambda$ is a tunable regularization hyperparameter (we use $\lambda =0.01$ as in \cite{Meinert2022DER}), $w_{St}$ is the width of the $t-$distribution, and $\Phi$ is the total evidence $\Phi = 2 \nu + \alpha$.
For a full derivation, see \cite{Meinert2022DER}, which we also summarize in Appendix \ref{ap:DER}.

We use the modified definitions of aleatoric uncertainty from \cite{Meinert2022DER}.
The aleatoric uncertainty is the width of the $t-$distribution, which resembles a normal distribution: $\sigma_{\rm al} \equiv w_{\rm St} =  \sqrt{\frac{\beta (1 + \nu)}{\alpha \nu}}$.

\begin{figure}
    \centering
    \includegraphics[scale=0.5, trim = 1cm 2cm 0cm 2.5cm]{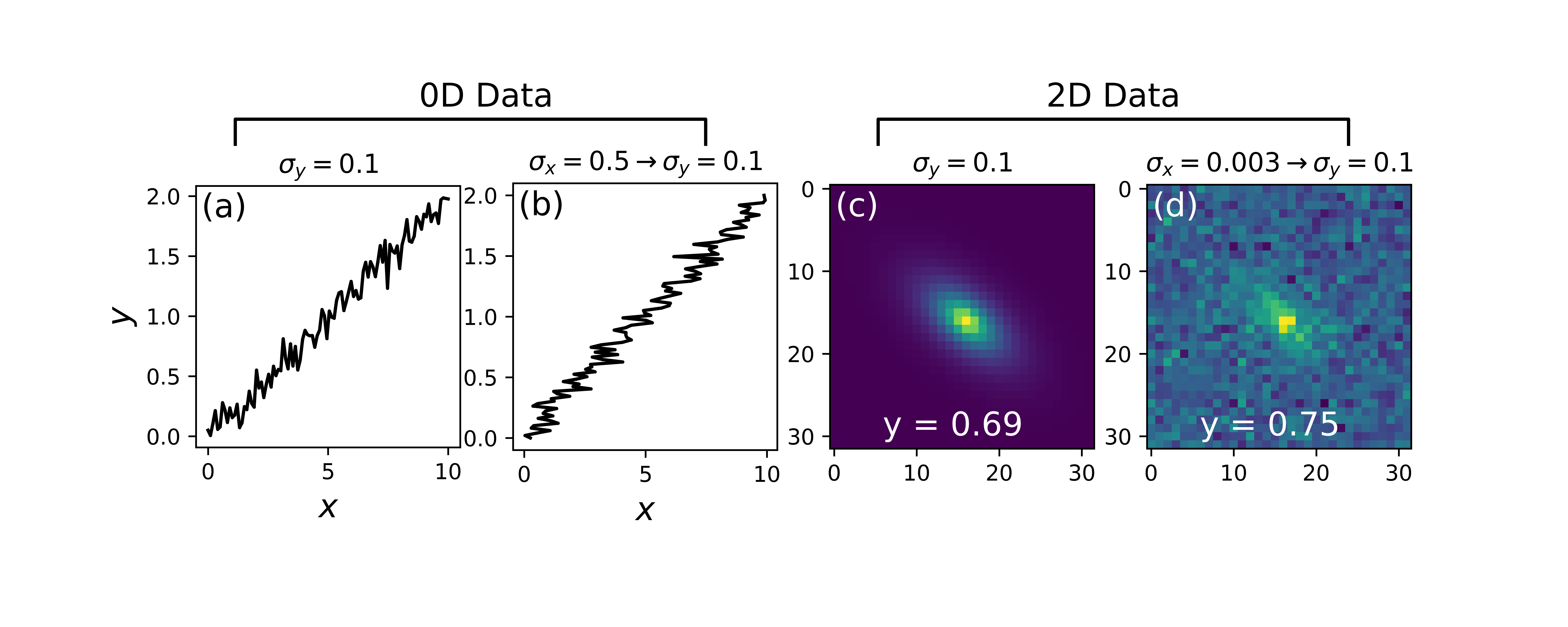}
    \caption{Data examples for the four experimental designs: a) output uncertainty for the 0D linear regression, b) input uncertainty for 0D, c) output uncertainty for the 2D imaging data, and d) input uncertainty for the 2D data. The noise level is high for all panels: $\sigma_{y} = 0.1$. For the case of the input uncertainty panels (b and d), the uncertainty is injected on the input variable, $\sigma_{x}$, and uncertainty propagation results in a $\sigma_{y}$ value of 0.1.}
    \label{fig:experimental_design}
\end{figure}

\noindent\textbf{Experimental Design:} 
Figure \ref{fig:experimental_design} illustrates the experimental setup for an example of high-noise data.

The 0D data are from a simple linear regression model: $y = mx $. 
The values of $x$ are linearly spaced between 0 and 10.
The data are designed so that the $y$ distribution is uniform, $\mathcal{U}[0,2]$.
The training/validation/test set size is 90k/10k/10k.
To create data for the output uncertainty experiments, we inject noise directly on the prediction or label, such that $y_{\rm noisy} = y + \mathcal{N}(0,\sigma_y^2)$.
The models are trained on $(x,y_{\rm noisy})$ pairs.
In the input uncertainty experiments, we inject noise via the input variable $x_{\rm noisy} = x + \mathcal{N}(0,\sigma_x^2)$ and the models are trained on $(x_{\rm noisy},y)$ pairs.

For the 2D data, we use the software package \href{https://github.com/deepskies/DeepBench}{\texttt{DeepBench}} (\cite{DeepBench} in prep) to generate $32\times32$-pixel galaxy images by varying the S\'ersic radius, amplitude, and position angle within ranges $[0, 0.01]$, $[1, 10]$, and $[-1.5, 1.5]$, respectively.
We are motivated by real-world uncertainty examples in astronomical imaging to generate a 2D dataset in addition to the 0D tabular dataset.
The output variable $y$ is the sum of the pixel values.
The dataset is designed to be uniform in $y$ over a range $[0,2]$ for a training/validation/test set size of 4500/500/500.
For the output uncertainty experiments, we inject a random normal distribution with mean zero and standard deviation $\sigma_y$ directly on $y$.
For the input uncertainty experiments, we inject a random normal distribution with mean zero and standard deviation $\sigma_x$ on each pixel, which results in a normal distribution in $\sigma_y$ after propagation.
We use a random normal distribution because the DE and the DER methods assume that the output variable is distributed as a random normal distribution.

We distinguish between the predicted aleatoric uncertainty from the methods, $\sigma_{\rm al}$, which is measured as an uncertainty on the output variable, and the true uncertainty on the output variable, $\sigma_y$.
The true uncertainty on the output variable is either directly known in the case of the output uncertainty experiments or is known through uncertainty propagation for the input uncertainty experiments.
The true output uncertainty has low, medium, and high values: $\sigma_y \in [0.01, 0.05, 0.1]$.
These noise levels are chosen such that the high noise value datasets have an uncertainty level that is on average $10\%$ of the output variable $y$.
For input uncertainty, we inject noise $\sigma_x$ and calculate the expected $\sigma_y$ uncertainty value via standard rules of error propagation described in Appendix \ref{ap:error_prop}.
The relationship between $\sigma_y$ and $\sigma_x$ for the 0D data is $\sigma_y = |m|\sigma_x$, where $m$ is the slope of the line, and the relationship for the 2D data is $\sigma_y = 32\sigma_x$.

\noindent\textbf{UQ Model Architectures:}
We use the \href{https://pypi.org/project/deepuq/}{\texttt{DeepUQ}} package to define the model architecture and perform our experiments.
We also present the \href{https://github.com/deepskies/DeepUQ-neurIPS-WS-2024}{\texttt{DeepUQ-neurIPS-WS-2024}} repository as an accompaniment to the paper, with notebook examples of how to reproduce the exact models, figures, and tables in this paper.
Both UQ methods use the same fully connected layer network architecture, which is two hidden layers of 64 neurons each.
The hidden layer neurons utilize a ReLu activation function. 
For the 0D experiments, the networks use two input neurons (the $m$ value and the $x$ value for a single point), while the 2D experiments use the $32\times32$ pixel input.
We use five convolutional layers for the 2D networks before the fully connected layers: the architecture is a series of convolutional filters that increase in depth and decrease in size further into the neural network.
These layers are interspersed with 2D pooling.
For all models, we use the Adam optimizer with an initial learning rate of 0.001.

For the DE method, two output neurons correspond to $\mu$ and $\sigma^2$, such that $y \sim \mathcal{N} (\mu, \sigma^2)$.
For the DER method, four output neurons correspond to the parameters $\gamma$, $\nu$, $\alpha$, and $\beta$. 
The output neurons utilize a softplus activation if a positive value is required (i.e., for $\sigma^2$ for the DE, and for $\nu$, $\alpha$, and $\beta$ for the DER) and a linear activation for all other outputs (i.e., for $\mu$ for the DE, and $\gamma$ for the DER).
For more details of the software package \texttt{DeepUQ}, see Appendix \ref{ap:DeepUQ}.

\section{Results}
\label{sec:data}

\begin{figure}
    \centering
    \includegraphics[scale=0.26, trim = 0cm 1.25cm 0cm 0cm]{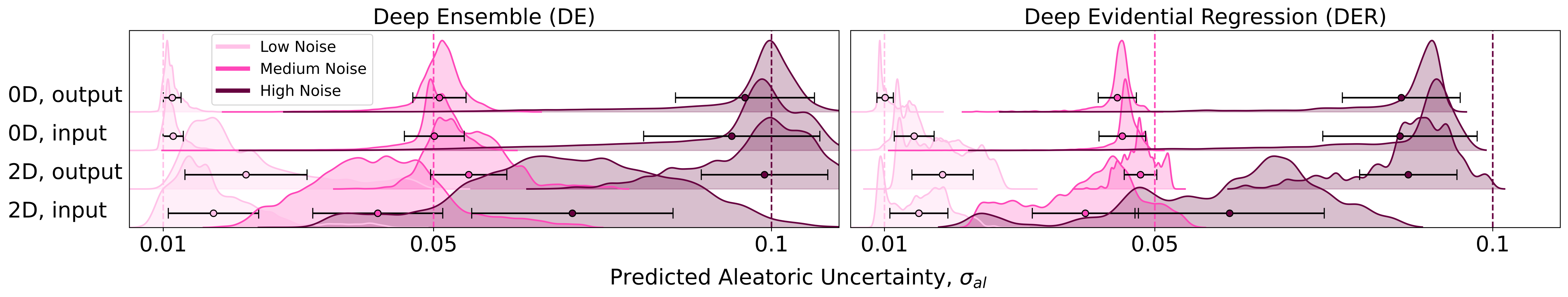}
    \caption{Distribution of predicted $\sigma_{\rm al}$ values. The circular point for each distribution is the sample mean and the black error bars show the $\rm {std}(\sigma_{\rm al})$ confidence range, the standard deviation of the $\sigma_{\rm al}$ distribution. The vertical dashed lines demonstrate the true output uncertainty values, $\sigma_y$, which vary by noise level. The light pink, medium pink, and purple distributions correspond to the low-, medium-, and high-noise models.}
    \label{fig:sigma_in_sigma_out}
\end{figure}

We run both UQ methods for all four experimental setups and all three noise levels and find that all models converge, with final mean-square error (MSE) values at epoch 99 ranging from 0.0001 to 0.01 (Appendix \ref{ap:model_diagnostics}).
Furthermore, the DE and DER methods have comparable final MSE values for each noise level, and the NIG loss and $\beta-$NLL loss values are similar for each UQ method across experiments for a fixed noise level.
This indicates that the different methods of uncertainty injection and the different data dimensionalities are all equally adequately learning to predict the relationship between input and output values.
To test desideratum (i), we display the distribution of predicted aleatoric uncertainties for the test set for the different noise levels in Figure \ref{fig:sigma_in_sigma_out}.
We use the standard deviation of the predicted uncertainty values, $\rm{std}(\sigma_{\rm al})$, to assess desiderata (ii) and (iii): whether the predicted uncertainty value $\sigma_{\rm al}$ is consistent with the true value $\sigma_y$ for all four experiments and all three noise levels.

\section{Discussion}

The predicted aleatoric uncertainty increases proportionately with the true injected uncertainty.
The models are sensitive to the true uncertainty, which bolsters confidence in these UQ methods.
It further confirms the findings of \cite{Egele2021arXiv211013511E}, where the automated Deep Ensemble method (AutoDEUQ) produces predicted aleatoric uncertainty that scales with uncertainty injected on the output variable.
Additionally, \cite{Bramlageinproceedings} find that the predicted aleatoric uncertainty from DER models increases for 0D and 2D experiments where uncertainty is increased on both the input and output variables.

When we require that the predicted uncertainty falls within $\rm{std}(\sigma_{\rm al})$ of the true uncertainty to be considered `well calibrated', we find that only seven of the twelve DE experiments and two of the twelve of the DER experiments satisfy this requirement.
Furthermore, for both methods, the degree of miscalibration depends on the experiment's dimensionality and the type of uncertainty injection.
Desiderata (ii) and (iii) are therefore violated for both experiments.

For the DE method, the 0D experiments are calibrated for the medium- and high-noise models (Figure \ref{fig:sigma_in_sigma_out}, left).
Both of the 0D low-noise experiments slightly overestimate the uncertainty.
Overall, the uncertainty estimates are the least calibrated for the most complex experimental setup (input, 2D; bottom row).
For the DER method (Figure \ref{fig:sigma_in_sigma_out}, right), the majority of experiments across all noise levels produce miscalibrated uncertainty estimates, over-estimating (low-noise) and under-estimating (medium- and high-noise) the predictive uncertainty. 
The exceptions are the 0D output low-noise model and the 2D output medium-noise model, which are calibrated.
For both methods, the most inaccurate experiment is the 2D input uncertainty experiment.
Within this experiment, the high-noise models are the least calibrated.

In \cite{Bramlageinproceedings}, the authors perform regression experiments for a DER model; the network is well calibrated for the 0D dataset but underestimates the true uncertainty for the 2D dataset injected with output uncertainty.
They suggest a recalibration step in more complex domains (i.e., 2D data) to ameliorate this concern, where recalibration involves training an auxiliary isotonic regression model so that the predicted uncertainties are calibrated to the cumulative density function of the data.
We expand upon the work of \cite{Bramlageinproceedings} to also include uncertainty injected on the input variable. 
For both types of uncertainty injection, we identify a miscalibration in aleatoric uncertainty for the DER models and identify that the effect is worse for high-dimensionality and high-noise experiments where the uncertainty injection is on the input variable.

\section{Conclusions and Outlook}
\label{sec:conclusion}
We explore aleatoric uncertainties predicted by two deep learning UQ approaches --- Deep Ensembles (DE) and Deep Evidential Regression (DER).
We compare the aleatoric uncertainty predictions of these two methods to the true uncertainty for four different experiments and three noise levels. 
Both methods meet desideratum (i): the aleatoric uncertainties scale with the injected uncertainty.

However, for our experiments, the methods both fail to meet desiderata (ii) and (iii), that the predicted aleatoric uncertainties be well-calibrated, i.e., consistent to $\rm{std}(\sigma_{\rm al})$ with the true uncertainties, and that they meet this requirement across all experiments.
Notably, most DER experiments underestimate the uncertainty for medium- and high-noise models and overestimate the uncertainty for low-noise models.
The DE experiments deviate mostly for the more complex 2D input uncertainty experiments.
The predicted uncertainties are the least accurate for both methods for the 2D, input uncertainty, and high-noise experiments.
While these observations do not imply inherent deficiencies in DE and DER, they highlight that further research would be beneficial to assess whether these methods require post-facto calibration, particularly for high-noise and high-dimensional settings.

Some limitations of our work are: Our conclusions apply only to our toy datasets.
We do not demonstrate performance of these methods on real world datasets with higher complexity.
Additionally, our conclusions apply only to homoskedastic noise generated from a Gaussian distribution; expanding this to non-Gaussian distributions and heteroskedastic noise is a direction for future work.

\bibliography{main}

\begin{thebibliography}{10}

\bibitem{Amini2019DER}
Alexander {Amini}, Wilko {Schwarting}, Ava {Soleimany}, and Daniela {Rus}.
\newblock {Deep Evidential Regression}.
\newblock {\em arXiv e-prints}, page arXiv:1910.02600, October 2019.

\bibitem{Angelopoulos2021arXiv210707511A}
Anastasios~N. {Angelopoulos} and Stephen {Bates}.
\newblock {A Gentle Introduction to Conformal Prediction and Distribution-Free Uncertainty Quantification}.
\newblock {\em arXiv e-prints}, page arXiv:2107.07511, July 2021.

\bibitem{Bramlageinproceedings}
Lennart Bramlage, Michelle Karg, and Cristóbal Curio.
\newblock Plausible uncertainties for human pose regression.
\newblock In {\em EEE/CVF International Conference on Computer Vision (ICCV)}, pages 15087--15096, 10 2023.

\bibitem{brando2022thesis}
A.~Brando.
\newblock {\em Aleatoric uncertainty modelling in regression problems using deep learning}.
\newblock PhD thesis, Universitat de Barcelona, 2022.

\bibitem{brando2023standardizing}
Axel Brando, Isabel Serra, Enrico Mezzetti, Francisco~Javier Cazorla~Almeida, and Jaume Abella~Ferrer.
\newblock Standardizing the probabilistic sources of uncertainty for the sake of safety deep learning.
\newblock In {\em Proceedings of the Workshop on Artificial Intelligence Safety 2023 (SafeAI 2023) co-located with the Thirty-Seventh AAAI Conference on Artificial Intelligence (AAAI 2023): Washington DC, USA, February 13-14, 2023.}, volume 3381. CEUR Workshop Proceedings, 2023.

\bibitem{Caldeira2020arXiv200410710C}
Jo{\~a}o {Caldeira} and Brian {Nord}.
\newblock {Deeply Uncertain: Comparing Methods of Uncertainty Quantification in Deep Learning Algorithms}.
\newblock {\em arXiv e-prints}, page arXiv:2004.10710, April 2020.

\bibitem{Chung2021arXiv210910254C}
Youngseog {Chung}, Ian {Char}, Han {Guo}, Jeff {Schneider}, and Willie {Neiswanger}.
\newblock {Uncertainty Toolbox: an Open-Source Library for Assessing, Visualizing, and Improving Uncertainty Quantification}.
\newblock {\em arXiv e-prints}, page arXiv:2109.10254, September 2021.

\bibitem{Egele2021arXiv211013511E}
Romain {Egele}, Romit {Maulik}, Krishnan {Raghavan}, Bethany {Lusch}, Isabelle {Guyon}, and Prasanna {Balaprakash}.
\newblock {AutoDEUQ: Automated Deep Ensemble with Uncertainty Quantification}.
\newblock {\em arXiv e-prints}, page arXiv:2110.13511, October 2021.

\bibitem{GalMCDropout2015arXiv150602142G}
Yarin {Gal} and Zoubin {Ghahramani}.
\newblock {Dropout as a Bayesian Approximation: Representing Model Uncertainty in Deep Learning}.
\newblock {\em arXiv e-prints}, page arXiv:1506.02142, June 2015.

\bibitem{gal2022bayesian}
Yarin Gal, Petros Koumoutsakos, François Lanusse, et~al.
\newblock Bayesian uncertainty quantification for machine-learned models in physics.
\newblock {\em Nature Reviews Physics}, 4:573--577, 2022.

\bibitem{Hullermeier2019arXiv191009457H}
Eyke {H{\"u}llermeier} and Willem {Waegeman}.
\newblock {Aleatoric and Epistemic Uncertainty in Machine Learning: An Introduction to Concepts and Methods}.
\newblock {\em arXiv e-prints}, page arXiv:1910.09457, October 2019.

\bibitem{Kendall2017arXiv170304977K}
Alex {Kendall} and Yarin {Gal}.
\newblock {What Uncertainties Do We Need in Bayesian Deep Learning for Computer Vision?}
\newblock {\em arXiv e-prints}, page arXiv:1703.04977, March 2017.

\bibitem{Ku2010NotesOT}
Harry~H. Ku.
\newblock Notes on the use of propagation of error formulas.
\newblock 2010.

\bibitem{Lakshminarayanan2016arXiv161201474L}
Balaji {Lakshminarayanan}, Alexander {Pritzel}, and Charles {Blundell}.
\newblock {Simple and Scalable Predictive Uncertainty Estimation using Deep Ensembles}.
\newblock {\em arXiv e-prints}, page arXiv:1612.01474, December 2016.

\bibitem{LV2000}
Jouko Lampinen and Aki Vehtari.
\newblock Bayesian techniques for neural networks — review and case studies.
\newblock In {\em 2000 10th European Signal Processing Conference}, pages 1--8, 2000.

\bibitem{Meinert2022DER}
Nis {Meinert}, Jakob {Gawlikowski}, and Alexander {Lavin}.
\newblock {The Unreasonable Effectiveness of Deep Evidential Regression}.
\newblock {\em arXiv e-prints}, page arXiv:2205.10060, May 2022.

\bibitem{Nix374138}
D.A. Nix and A.S. Weigend.
\newblock Estimating the mean and variance of the target probability distribution.
\newblock In {\em Proceedings of 1994 IEEE International Conference on Neural Networks (ICNN'94)}, volume~1, pages 55--60 vol.1, 1994.

\bibitem{Nixon2019arXiv190401685N}
Jeremy {Nixon}, Mike {Dusenberry}, Ghassen {Jerfel}, Timothy {Nguyen}, Jeremiah {Liu}, Linchuan {Zhang}, and Dustin {Tran}.
\newblock {Measuring Calibration in Deep Learning}.
\newblock {\em arXiv e-prints}, page arXiv:1904.01685, April 2019.

\bibitem{PS2017}
Nicholas~G. Polson and Vadim Sokolov.
\newblock {Deep Learning: A Bayesian Perspective}.
\newblock {\em Bayesian Analysis}, 12(4):1275 -- 1304, 2017.

\bibitem{Rodrigues2023MLS&T...4d5019R}
Nat{\'a}lia V.~N. {Rodrigues}, L.~{Raul Abramo}, and Nina S.~T. {Hirata}.
\newblock {The information of attribute uncertainties: what convolutional neural networks can learn about errors in input data}.
\newblock {\em Machine Learning: Science and Technology}, 4(4):045019, December 2023.

\bibitem{Scalia2019arXiv191003127S}
Gabriele {Scalia}, Colin~A. {Grambow}, Barbara {Pernici}, Yi-Pei {Li}, and William~H. {Green}.
\newblock {Evaluating Scalable Uncertainty Estimation Methods for DNN-Based Molecular Property Prediction}.
\newblock {\em arXiv e-prints}, page arXiv:1910.03127, October 2019.

\bibitem{Seitzer2022arXiv220309168S}
Maximilian {Seitzer}, Arash {Tavakoli}, Dimitrije {Antic}, and Georg {Martius}.
\newblock {On the Pitfalls of Heteroscedastic Uncertainty Estimation with Probabilistic Neural Networks}.
\newblock {\em arXiv e-prints}, page arXiv:2203.09168, March 2022.

\bibitem{Seitzer2022}
Maximilian {Seitzer}, Arash {Tavakoli}, Dimitrije {Antic}, and Georg {Martius}.
\newblock {On the Pitfalls of Heteroscedastic Uncertainty Estimation with Probabilistic Neural Networks}.
\newblock {\em arXiv e-prints}, page arXiv:2203.09168, March 2022.

\bibitem{T2004}
D.~M. Titterington.
\newblock {Bayesian Methods for Neural Networks and Related Models}.
\newblock {\em Statistical Science}, 19(1):128 -- 139, 2004.

\bibitem{Tran2019arXiv191210066T}
Kevin {Tran}, Willie {Neiswanger}, Junwoong {Yoon}, Qingyang {Zhang}, Eric {Xing}, and Zachary~W. {Ulissi}.
\newblock {Methods for comparing uncertainty quantifications for material property predictions}.
\newblock {\em arXiv e-prints}, page arXiv:1912.10066, December 2019.

\bibitem{Ulmer2021arXiv211003051U}
Dennis {Ulmer}, Christian {Hardmeier}, and Jes {Frellsen}.
\newblock {Prior and Posterior Networks: A Survey on Evidential Deep Learning Methods For Uncertainty Estimation}.
\newblock {\em arXiv e-prints}, page arXiv:2110.03051, October 2021.

\bibitem{Valdenegro-Toro2024arXiv240618787V}
Matias {Valdenegro-Toro}, Ivo~Pascal {de Jong}, and Marco {Zullich}.
\newblock {Unified Uncertainties: Combining Input, Data and Model Uncertainty into a Single Formulation}.
\newblock {\em arXiv e-prints}, page arXiv:2406.18787, June 2024.

\bibitem{DeepBench}
M.~{Voetberg}, Ashia {Livaudais}, Becky {Nevin}, Omari {Paul}, and Brian {Nord}.
\newblock Deepbench: A simulation package for physical benchmarking data.
\newblock Submitted to Journal of Open Source Software, 2024.
\newblock Manuscript submitted for publication.

\bibitem{Wright2000}
W.~Wright, Guillaume Ramage, Dan Cornford, and Ian Nabney.
\newblock Neural network modelling with input uncertainty: Theory and application.
\newblock {\em VLSI Signal Processing}, 26:169--188, 08 2000.

\end{thebibliography}
\bibliographystyle{plain}

\newpage
\appendix
\onecolumn

\begin{ack} 
\section{Funding}
\label{sec:app:funding}

\ackfundingdeepskies
\ackfundingdoegeneral
\ackfundingldrdnord

\section{Author Contributions}
\label{sec:app:funding}

\contribnevin
\contribciprijanovic
\contribnord

We thank the following colleagues for their insights and discussions during the development of this work: Sreevani Jarugula.

\end{ack}

\newpage

\section{Uncertainty propagation}
\label{ap:error_prop}

Here we describe our process for propagating uncertainty injected on the input variable $\sigma_x$ onto uncertainty on the output variable $\sigma_y$.
For a generic function $y = f(x_1, x_2, ... x_N)$, where $y$ is the dependent variable and $x_1$, $x_2$, and so on are the independent variables, the standard deviation of $y$, $\sigma_y$, can be written in terms of uncertainty on the $x$ variables \cite{Ku2010NotesOT}:

\begin{equation}
    \sigma_y = \sqrt{{(\frac{\partial f}{\partial x_1})}^2\sigma_{x_1}^2 + {(\frac{\partial f}{\partial x_2})}^2\sigma_{x_2}^2 + 2 (\frac{\partial f}{\partial x_1}) (\frac{\partial f}{\partial x_2}) \sigma_{x_1x_2}},
\end{equation}
where the final covariance term ($\sigma_{x_1x_2}$) can be dropped when the correlation between uncertainty terms is negligible, as is the case for both of our data dimensionalities. 

For the 0D linear regression case, $y=mx$, the partial derivative only exists relative to $x$, which has associated uncertainty, so this equation reduces to:
\begin{equation}
    \sigma_y = |m|\sigma_x.
    \label{eq:error_prop_linear}
\end{equation}

For the case of the 2D image noise injection, we inject a standard normal value for each pixel and calculate the predicted value $y$ as a sum of all pixel values.
The partial derivative terms are all equal to 1 because this is a summation. Since we inject the same value of $\sigma$ for all pixels, the formula then becomes:
\begin{equation}
    \sigma_y = \sqrt{\sum_{i=1}^N \sigma_{x_i}^2} = 32 \sigma_x
    \label{eq:error_prop_image_2}
\end{equation}
where $i$ is the index of all ($N$) pixels, and the images are $32\times32$ pixels.

\section{Deep Ensembles}
\label{ap:DE}
A common approach for quantifying aleatoric uncertainty in regression tasks with deep neural networks is to assume that the regression output $y$ follows a distribution and to predict the parameters of this distribution. 
One standard technique is to assume that the errors are heteroskedastic\footnote{The data we generate are homoskedastic, with a constant $\sigma^2$ for each noise level, while the Gaussian model in MVE is heteroskedastic, allowing the predicted $\sigma^2$ value to vary across points. These experiments, where we test MVE with homoskedastic data, do not break the model's assumption. Instead, we ensure that the model can still accurately predict constant uncertainty when the true distribution is homoskedastic. We are particularly interested here in the calibration of the uncertainty predictions; assessing the model's ability to return a distribution of uncertainty values is a compelling topic for future research.} and to model the distribution of $y$ as a Gaussian parameterized by mean $\mu$ and variance $\sigma^2$, where the predicted values $y_i \sim \mathcal{N}(\mu(x_i), \sigma^2(x_i))$. 
The model is trained using maximum likelihood estimation by minimizing the negative log likelihood loss under the training set distribution $p(X,Y)$:

\begin{equation}
    \begin{split}
        \mathcal{L}_{\rm NLL} &= -\mathrm{log}\ p(Y|X)   \\
            &= \frac{1}{N} \sum_{i=0}^N \left[ \frac{1}{2} \mathrm{log}\ \sigma^2(x_i) + \frac{(y_i - \mu(x_i))^2}{2\sigma^2(x_i)} + \mathrm{C} \right],
    \end{split}
\end{equation}
where $\mu(x_i)$ and $\sigma^2(x_i)$ are the model outputs for each training data point using the model with the optimal set of internal parameters.
This technique is known as mean-variance estimation \citep[MVE;][]{Nix374138,Kendall2017arXiv170304977K}.

We use a modified loss function for training, known as the $\beta$-NLL loss.
This loss is proposed by \cite{Seitzer2022} as a means for avoiding a commonly observed problem in MVEs, where the variance artificially enlarges resulting in a poor estimate of the mean.
The $\beta$ parameter helps ensure convergence of the network predictions for $\mu(x_i)$ and $\sigma^2(x_i)$:

\begin{equation}
  \mathcal{L}_{\beta-\rm NLL} = 
       \frac{1}{N} \sum_{i=0}^N 
       \left[ \sigma^{2\beta}(x_i)\left[\frac{1}{2}  \mathrm{log}\ \sigma^2(x_i) + 
        \frac{(y_i - \mu(x_i)) ^2}{2\sigma^2(x_i)} 
        + 
        \mathrm{C}\right] \right],
\end{equation}

where the contribution of each data point is weighted by its $\beta$-exponentiated variance estimate.
This modified loss simplifies to the standard Gaussian negative log likelihood for $\beta=0$ and the mean-squared error (MSE) loss for $\beta=1$.
We experiment with several prescriptions for $\beta$ including constant values $\beta = 0.0, 0.5, 1.0$ and several situations where $\beta$ changes throughout training, including a linearly decreasing $\beta$ value, 1 to 0 and two step functions, where $\beta$ decreases from 1 to 0.5 and 1 to 0.0 at half the total number of epochs.
We ultimately select a $\beta$ value of 0.5, which is recommended by \cite{Seitzer2022}.

\section{Deep Evidential Regression}
\label{ap:DER}
Similar to MVE, we assume the training data are drawn from a Gaussian likelihood distribution $y_i\sim \mathcal{N}(\mu(x_i), \sigma^2(x_i))$.
We also place a Gaussian prior on the mean $\mu$ and an Inverse-Gamma prior on the variance $\sigma^2$:

\begin{equation}
    \begin{split}
        \mu_j &\sim \mathcal{N}(\gamma_j, \sigma^2_j/\nu_j), \\
        \sigma^2_j &\sim \Gamma^{-1}(\alpha_j, \beta_j),
    \end{split}
\end{equation}
where $j$ is a sample drawn from these hyperprior distributions; $\Gamma(\cdot)$ is the gamma function; and ${\bf m} = (\gamma, \nu, \alpha, \beta)$ are hyperparameters of these distributions, where $\gamma \in \mathbb{R}$, $\nu > 0$, $\alpha > 1$, and $\beta > 0$.

One can then formulate the conjugate prior distribution as a Normal-Inverse-Gamma (NIG) distribution; for a full derivation, see \cite{Amini2019DER}.
Drawing a sample from the NIG distribution yields a single instance $j$ of the likelihood function: the NIG hyperparameters ${\bf m}$ control the location and dispersion (uncertainty) of the likelihood function $\mathcal{N}(\mu_j,\sigma_j^2)$.
We will later use the hyperparameters of this higher-order evidential distribution to define the aleatoric uncertainty; these higher-order parameters determine the lower-order likelihood distribution from which observations are drawn.

To fit the model, we define a marginal likelihood $p(y_i|\textbf{m})$, which is done using Bayesian probability theory in \cite{Amini2019DER}: the conjugate prior defined above is combined with the Gaussian likelihood and integrated over the parameters $\mu$ and $\sigma^2$.
An analytic solution to the marginal likelihood is the Student $t-$distribution:

\begin{equation}
    L_{i, \rm NIG} = {\rm St}_{2\alpha}(y_i|\gamma,\frac{\beta(1+\nu)}{\nu\alpha}),
\end{equation}
where ${\rm St}_{2\alpha}$ is a $t-$distribution with $2\alpha$ degrees of freedom.

The negative log-likelihood loss of this distribution and the addition of an additional term weighted by the width of the $t-$distribution provides the $\mathcal{L}_{\rm NIG}$ loss that we use for training:
$\mathcal{L}_{\rm NIG} = \frac{1}{N} \sum_{i=1}^N \left[- {\rm log}\  L_i^{\rm NIG} + \lambda |\frac{y_i-\gamma}{w_{\rm St}}| \Phi \right]$.

\section{Software package \texttt{DeepUQ}}
\label{ap:DeepUQ}

\href{https://pypi.org/project/deepuq/}{\texttt{DeepUQ}} is a software package that provides modules, utilities, and scripts for setting the hyperparameters and training both DE and DER models and analyzing the predicted aleatoric uncertainties.
It is also designed to be tunable to insert additional UQ methods and/or to create additional noise profiles for uncertainty injection on the 0D or 2D data.

\texttt{DeepUQ} provides the following modules and scripts:
\begin{itemize}
    \item \texttt{data.py} - A module for generating data with accompanying controllable injected aleatoric uncertainty on either the input or output variables.
    \item \texttt{model.py} - A module that provides tunable loss functions, network architecture, and hyperparameters for the DE and DER methods.
    \item \texttt{train.py} - A module for the training procedure for both models.
    \item \texttt{DeepEnsemble.py} and \texttt{DeepEvidentialRegression.py} - Scripts for generating data, initializing the methods, and training the models.
\end{itemize} 

The \href{https://github.com/deepskies/DeepUQ-neurIPS-WS-2024}{\texttt{DeepUQ-neurIPS-WS-2024}} repository provides additional details for how run the \texttt{DeepUQ} scripts to exactly reproduce the results of the paper, including the models, figures, and tables.

\section{Model Loss}
\label{ap:model_diagnostics}

We report the values for the MSE metric and the NIG or $\beta-$NLL loss for the validation set for the low- and high-noise models for the final epoch of training in Tables \ref{tab:losslow} and \ref{tab:losshigh}, respectively.

\begin{table}[ht]
    \centering
    \begin{tabular}{|c|c|c|c|c|c|c|c|c|}
        \hline
        \multirow{2}{*}{Metric} & \multicolumn{4}{c|}{0D Data} & \multicolumn{4}{c|}{2D Data} \\
        \cline{2-9}
        & \multicolumn{2}{c|}{Output Injection} & \multicolumn{2}{c|}{Input Injection} & \multicolumn{2}{c|}{Output Injection} & \multicolumn{2}{c|}{Input Injection} \\
        \cline{2-9}
        & DE & DER & DE & DER & DE & DER & DE & DER \\
        \hline
        MSE Metric & 0.0001 & 0.0001& 0.0001 &  0.0001& 0.0006 & 0.0002& 0.0002 & 0.0001 \\
        \hline
        Loss & -0.0502 & -3.0890 & -0.0416 & -2.9338 & -0.0426 & -2.7691 & -0.0993 & -3.0639 \\
        \hline
    \end{tabular}
    \caption{Loss values for the final epoch of the low-noise experiments. We provide the mean-square error (MSE) metric and $\beta-$NLL/NIG loss for DE/DER methods in 0D and 2D for uncertainty injected on the output and input variables.}
    \label{tab:losslow}
\end{table}

\begin{table}[ht]
    \centering
    \begin{tabular}{|c|c|c|c|c|c|c|c|c|}
        \hline
        \multirow{2}{*}{Metric} & \multicolumn{4}{c|}{0D Data} & \multicolumn{4}{c|}{2D Data} \\
        \cline{2-9}
        & \multicolumn{2}{c|}{Output Injection} & \multicolumn{2}{c|}{Input Injection} & \multicolumn{2}{c|}{Output Injection} & \multicolumn{2}{c|}{Input Injection} \\
        \cline{2-9}
        & DE & DER & DE & DER & DE & DER & DE & DER \\
        \hline
        MSE Metric &0.0098  & 0.0097 &0.0091  & 0.0092 & 0.0099  & 0.0086 & 0.0047 & 0.0042  \\
        \hline
        Loss &  -0.1724 & -0.8728 & -0.1678 & -0.9018 &  -0.1767 & -0.9202 & -0.1330 &  -1.3358\\
        \hline
    \end{tabular}
    \caption{Loss values for the final epoch of the high-noise experiments. We provide the mean-square error (MSE) metric and $\beta-$NLL/NIG loss for DE/DER methods in 0D and 2D for uncertainty injected on the output and input variables.}
    \label{tab:losshigh}
\end{table}

\end{document}